# A Robust Fuzzy Clustering Technique with Spatial Neighborhood Information for Effective Medical Image Segmentation

S.Zulaikha Beevi [1], M.Mohammed Sathik [2], K.Senthamaraikannan [3]

[1] Assistant Professor, Department of IT, National College of Engineering, Tamilnadu, India.

[2] Associate Professor, Department of Computer Science, Sathakathullah Appa College,Tamilndu, India.

[3] Professor & Head, Department of Statistics, Manonmaniam Sundaranar University, Tamilnadu, India.

*Abstract*-Medical image segmentation demands an efficient and robust segmentation algorithm against noise. The conventional fuzzy c-means algorithm is an efficient clustering algorithm that is used in medical image segmentation. But FCM is highly vulnerable to noise since it uses only intensity values for clustering the images. This paper aims to develop a novel and efficient fuzzy c-means clustering algorithm which is robust to noise. The proposed clustering algorithm uses fuzzy spatial information to calculate membership value. The input image is clustered using proposed ISFCM algorithm. A comparative study has been made between the conventional FCM and proposed ISFCM. The proposed approach is found to be outperforming the conventional FCM.

## I.INTRODUCTION

Data clustering is a common technique for statistical data analysis, which is used in many fields, including machine learning, data mining, pattern recognition, image analysis and bioinformatics. Clustering is the classification of similar objects into different groups, or more precisely, the partitioning of a data set into subsets (clusters), so that the data in each subset (ideally) share some common trait - often proximity according to some defined distance measure. Medical imaging techniques such as X - ray, computed tomography (CT), magnetic resonance imaging (MRI), positron emission tomography (PET), ultrasound (USG), etc. are indispensable for the precise analysis of various medical pathologies. Computer power and medical scanner data alone are not enough. We need the art to extract the necessary boundaries, surfaces, and segmented volumes these organs in the spatial and temporal domains. This art of organ extraction is segmentation. Image segmentation is essentially a process of pixel classification, wherein the image pixels are segmented into subsets by assigning the individual pixels to classes. These segmented organs and their boundaries are very critical in the quantification process for physicians and medical surgeons, in any branch of medicine, which deals with imaging [1]. Recently, fuzzy techniques are often applied as complementary to existing techniques and can contribute to the development of better and more robust methods, as it has been illustrated in numerous scientific branches. It seems to be proved that applications of fuzzy techniques are very successful in the area of image processing [2]. Moreover, the field of medicine has become a very attractive domain for the application of fuzzy set theory. This is due to the large role imprecision and uncertainty plays in the field [3]. The main objective of medical image segmentation is to extract and characterize anatomical structures with respect to some input features or expert knowledge. Segmentation methods that includes the classification of tissues in medical imagery can be performed using a variety of techniques. Many clustering strategies have been used, such as the crisp clustering scheme and the fuzzy clustering scheme, each of which has its own special characteristics [3]. The conventional crisp clustering method restricts each point of the data set to exclusively just one cluster. However, in many real situations, for images, issues such as limited spatial resolution, poor contrast, overlapping intensities, noise and intensity in homogeneities variation make this hard (crisp) segmentation a difficult task. The fuzzy set theory [4], which involves the idea of partial membership described by a membership function, fuzzy clustering as a soft segmentation method has been widely studied and successfully applied to image segmentation [5–11]. Among the fuzzy clustering methods, fuzzy *c*-means (FCM) algorithm [1] is the most popular method used in image segmentation because it has robust characteristics for ambiguity and can retain much more information than hard segmentation methods [5, 6].Although the conventional FCM algorithm works well on most noise-free images, it has a serious limitation: it does not incorporate any information about spatial context, which cause it to be sensitive to noise and imaging artifacts. In this paper, Improved Spatial FCM (ISFCM) clustering algorithm for image segmentation is proposed. The algorithm is developed by incorporating the spatial neighborhood information into the standard FCM clustering algorithm by *a priori* probability. The probability is given to indicate the spatial influence of the neighboring pixels on the centre pixel, which can be automatically decided in the implementation of the algorithm by the fuzzy membership. The new fuzzy membership of the current centre pixel is then recalculated with this probability obtained from above. The algorithm is initialized by a given histogram based FCM algorithm, which helps to speed up the convergence of the algorithm. Experimental results with medical images that the proposed method can achieve comparable results to those from many derivatives of FCM algorithm, that gives the method presented in this paper is effective.

## II. RELATED WORKS

There are huge amount of works related to enhancing the conventional FCM and other forms for image segmentation are found in the literature. Let us review some of them. Smaine Mazouzi and Mohamed Batouche [13] have presented an approach for improving range image segmentation, based on fuzzy regularization of the detected edges. Initially, a degraded version of the segmentation was produced by a new region growing- based algorithm. Next, the resulting segmentation was refined by a robust fuzzy classification of the pixels on the resulting edges which correspond to border of the extracted regions. Pixels on the boundary between two adjacent regions are labeled taking into account the two regions as fuzzy sets in the fuzzy classification stage, using an improved version of the Fuzzy C-Mean algorithm. The process was repeated for all region boundaries in the image. A two-dimensional fuzzy C-means (2DFCM) algorithm was proposed by Jinhua Yu and Yuanyuan Wang [14] for the molecular image segmentation. The 2DFCM algorithm was composed of three stages. The first stage was the noise suppression by utilizing a method combining a Gaussian noise filter and anisotropic diffusion techniques. The second stage was the texture energy characterization using a Gabor wavelet method. The third stage was introducing spatial constraints provided by the denoising data and the textural information into the two-dimensional fuzzy clustering. The incorporation of intensity and textural information allows the 2DFCM algorithm to produce satisfactory segmentation results for images corrupted by noise (outliers) and intensity variations. Hadi Sadoghi Yazdi and Jalal A. Nasiri [15] have presented a fuzzy image segmentation algorithm. In their algorithm, human knowledge was used in clustering features for fuzzy image segmentation. In fuzzy clustering, the membership values of extracted features for each pixel at each cluster change proportional to zonal mean of membership values and gradient mean of adjacent pixels. The direction of membership variations are specified using human interaction. Their segmentation approach was applied for segmentation of texture and documentation images and the results have shown that the human interaction eventuates to clarification of texture and reduction of noise in segmented images.G.Sudhavani and Dr.K.Sathyaprasad [16] have described the application of a modified fuzzy C-means clustering algorithm to the lip segmentation problem. The modified fuzzy C-means algorithm was able to take the initial membership function from the spatially connected neighboring pixels. Successful segmentation of lip images was possible with their method. Comparative study of their modified fuzzy C-means was done with the traditional fuzzy C-means algorithm by using Pratt's Figure of Merit. (2009) B.Sowmya and B.Sheelarani [9] have explained the task of segmenting any given color image using soft computing techniques. The most basic attribute for segmentation was image luminance amplitude for a monochrome image and color components for a color image. Since there are more than 16 million colors available in any given image and it was difficult to analyze the image on all of its colors, the likely colors are grouped together by image segmentation. For that purpose soft computing techniques have been used. The soft computing techniques used are Fuzzy C-Means algorithm (FCM), Possibilistic C - Means algorithm (PCM) and competitive neural network. A self estimation algorithm was developed for determining the number of clusters. Agus Zainal Arifin and Akira Asano [12] have proposed a method of image thresholding by using cluster organization from the histogram of an image. A similarity measure proposed by them was based on inter-class variance of the clusters to be merged and the intra-class variance of the new merged cluster. Experiments on practical images have illustrated the effectiveness of their method. (2006) An high speed parallel fuzzy C means algorithm for brain tumor image segmentation is presented by S. Murugavalli and V. Rajamani [17]. Their algorithm has the advantages of both the sequential FCM and parallel FCM for the clustering process in the segmentation techniques and the algorithm was very fast when the image size was large and also it requires less execution time. They have also achieved less processing speed and minimizing the need for accessing secondary storage compared to the previous results. The reduction in the computation time was primarily due to the selection of actual cluster centre and the accessing minimum secondary storage. (2006) T. Bala Ganesan and R. Sukanesh [18] have deals with Brain Magnetic Resonance Image Segmentation. Any medical image of human being consists of distinct regions and these regions could be represented by wavelet coefficients. Classification of these features was performed using Fuzzy Clustering method (FCM Fuzzy C-Means Algorithm). Edge detection technique was used to detect the edges of the given images. Silhouette method was used to find the strength of clusters. Finally, the different regions of the images are demarcated and color coded. (2008) H. C. Sateesh Kumar et al. [19] have proposed Automatic Image Segmentation using Wavelets (AISWT) to make segmentation fast and simpler. The approximation band of image Discrete Wavelet Transform was considered for segmentation which contains significant information of the input image. The Histogram based algorithm was used to obtain the number of regions and the initial parameters like mean, variance and mixing factor. The final parameters are obtained by using the Expectation and Maximization algorithm. The segmentation of the approximation coefficients was determined by Maximum Likelihood function. Histogram specification was proposed by Gabriel Thomas [20] as a way to improve image segmentation. Specification of the final histogram was done relatively easy and all it takes is the definition of a low pass filter and the amplification and attenuation of the peaks and valleys respectively or the standard deviation of the assumed Gaussian modes in the final specification. Examples showing better segmentation were presented. The attractive side of their approach was the easy implementation that was needed to obtain considerable better results during the segmentation process.

### III. PROPOSED IMPROVED SPATIAL FUZZY C-MEANS CLUSTERING (ISFCM)

#### A. Conventional FCM

Clustering is the process of finding groups in unlabeled dataset based on a similarity measure between the data patterns (elements) [12]. A cluster contains similar patterns placed together. The fuzzy clustering technique generates fuzzy partitions of the data instead of hard partitions. Therefore, data patterns may belong to several clusters, having different membership values with different clusters. The membership value of a data pattern to a cluster denotes similarity between the given data pattern to the cluster. Given a set of n data patterns, $X = x_1,...,x_k,...,x_n$, the fuzzy clustering technique minimizes the objective function, $O(U,C)$ :

$$O_{fcm}(U,C) = \sum_{k=1}^{n} \sum_{i=1}^{v} \left(u_{ik}\right)^m d^2(x_k, c_i)$$

(1)

where $x_k$ is the k-th D-dimensional data vector, $c_i$ is the center of cluster i, $u_{ik}$ is the degree of membership of $x_k$ in the i-th cluster, m is the weighting exponent, d $(x_k, c_i)$ is the distance between data $x_k$ and cluster center $c_i$, n is the number of data patterns, v is the number of clusters. The minimization of objective function J (U, C) can be brought by an iterative process in which updating of degree of membership $u_{ik}$ and the cluster centers are done for each iteration.

$$u_{ik} = \frac{1}{\sum_{J=1}^{V} \left(\frac{d_{ik}}{d_{jk}}\right)^{\frac{1}{m-1}}}$$

(2)

$$c_i = \frac{\sum_{k=1}^{n} \left(u_{ik}\right)^m x_k}{\sum_{k=1}^{n} \left(u_{ik}\right)^m}$$

(3)

where $\forall i$ $u_{ik}$ satisfies: $u_{ik} \in [0,1]$, $\forall k \sum_{i=1}^{v} u_{ik} = 1$ and

$0 < \sum_{k=1}^{n} u_{ik} < n$

Thus the conventional clustering technique clusters an image data only with the intensity values but it does not use the spatial information of the given image.

#### B. Initialization

The theory of Markov random field says that pixels in the image mostly belong to the same cluster as their neighbors. The incorporation of spatial information in the clustering process makes the algorithm robust to noise and blurred edges. But when using spatial information in the clustering optimization function may converge in local minima, so to avoid this problem the fuzzy spatial c means algorithm is initialized with the Histogram based fuzzy c-means algorithm. The optimization function for histogram based fuzzy clustering is given in the equation 4.

$$O_{hfcm}(U,C) = \sum_{l=1}^{L}\sum_{i=1}^{v}\left(u_{il}\right)^{m} H(l)\, d^{2}(l,c_{i}) \qquad (4)$$

where H is the histogram of the image of L-gray levels. Gray level of all the pixels in the image lies in the new discrete set G= {0,1,…,L-1}. The computation of membership degrees of H($l$) pixels is reduced to that of only one pixel with $l$ as grey level value. The member ship function $u_{il}$ and center for histogram based fuzzy c-means clustering can be calculated as.

$$u_{il} = \frac{1}{\sum_{J=1}^{V}\left(\dfrac{d_{li}}{d_{lj}}\right)^{\frac{1}{m-1}}} \qquad (5)$$

$$c_{i} = \frac{\sum_{l=1}^{L}\left(u_{il}\right)^{m} H(l)\, l}{\sum_{l=1}^{L}\left(u_{il}\right)^{m}} \qquad (6)$$

where $d_{li}$ is the distance between the center $i$ and the gray level $l$

*C. Proposed ISFCM*

The histogram based FCM algorithm converges quickly since it clusters the histogram instead of the whole image. The center and member ship values of all the pixels are given as input to the fuzzy spatial c-means algorithm. The main goal of the FSCM is to use the spatial information to decide the class of a pixel in the image.

The objective function of the proposed ISFCM is given by

$$O_{fcm}(U,C) = \sum_{k=1}^{n}\sum_{i=1}^{v}\left(u_{ik}^{s}\right)^{m} d^{2}(x_{k},c_{i}) \qquad (7)$$

$$u_{ik}^{s} = \frac{P_{ik}}{\left(\sum_{J=1}^{V}\left(\dfrac{d_{ik}}{d_{jk}}\right)^{\frac{1}{m-1}}\right)\left(N_{k}\sum_{z=1}^{N_{k}}\left(\sum_{J=1}^{V}\left(\dfrac{d_{iz}}{d_{jz}}\right)^{\frac{1}{m-1}}\right)\right)} \qquad (8)$$

The spatial membership function $u_{ik}^{s}$ of the proposed ISFCM is calculated using the equation (8).

where $P_{ik}$ is the apriori probability that k[th] pixel belongs to i[th] cluster and calculated as

$$P_{ik} = \frac{NN_{i}(k)}{N_{k}} \qquad (9)$$

where $NN_{i}(k)$ is the number of pixels in the neighborhood of k[th] pixel that belongs to cluster i after defuzzification. $N_{k}$ is the total number of pixels in the neighborhood. $d_{iz}$ is the distance between i[th] cluster and z[th] neighborhood of i[th] Thus the center $c_{i}^{s}$ of each cluster is calculated as

$$c_{i}^{s} = \frac{\sum_{k=1}^{n}\left(u_{ik}^{s}\right)^{m} x_{k}}{\sum_{k=1}^{n}\left(u_{ik}^{s}\right)^{m}} \qquad (10)$$

Two kinds of spatial information are incorporated in the member ship function of conventional FCM. Apriori probability and Fuzzy spatial information

*Apriori probability:* This parameter assigns a noise pixel to one of the clusters to which its neighborhood pixels belong. The noise pixel is included in the cluster whose members are majority in the pixels neighborhood.

*Fuzzy spatial information:* In the equation (8) the second term in the denominator is the average of fuzzy membership of the neighborhood pixel to a cluster. Thus a pixel gets higher membership value when their neighborhood pixels have high membership value with the corresponding cluster.

IV. RESULTS AND DISCUSSION

The proposed ISFCM algorithm converges very quickly because it gets initial parameters form already converged histogram based FCM. The proposed approach is applied on three kinds of images real world images, synthetic brain MRI image, original brain MRI image. In all the images additive Gaussian white noise is added with noise percentage level 0%, 5%, 10%, and 15% and corresponding results are shown. The quality of segmentation of the proposed algorithm(IFSCM) can be calculated by segmentation accuracy $A_{s}$ given as.

$$A_S = \frac{N_c}{T_p} \times 100 \qquad (11)$$

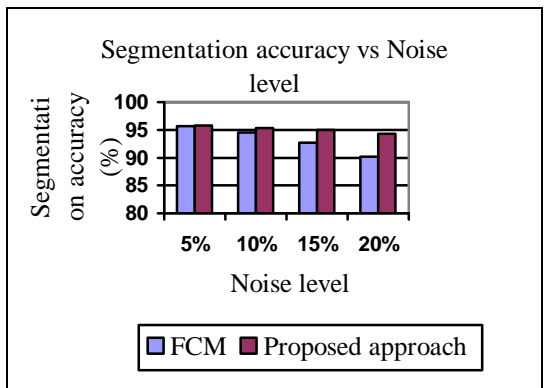

Fig.1. Segmentation accuracy of FCM, proposed approach without denoising in segmenting synthetic brain MRI images with different noise level.

$N_c$ is the number of correctly classified pixels and $T_p$ is the total is the total number pixels in the given image. Segmentation accuracy of FCM, proposed approach without denoising in segmenting synthetic brain MRI images with different noise level is shown in figure 1.

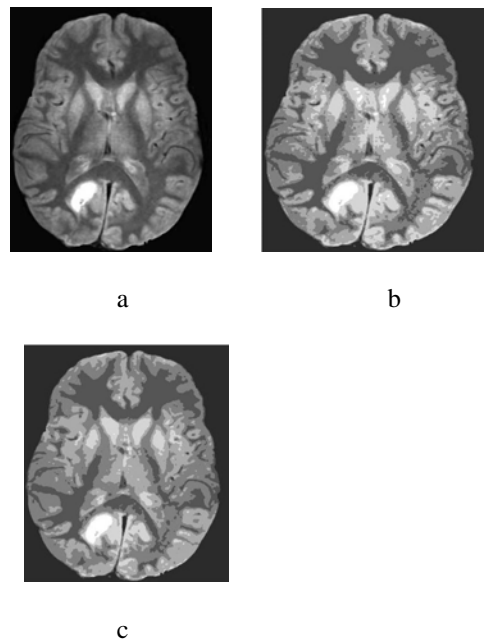

Fig.3. Segmentation results of original brain MRI image. (a) Original brain MRI image with tumor. (b) FCM segmentation result (c) Proposed approach without denoising

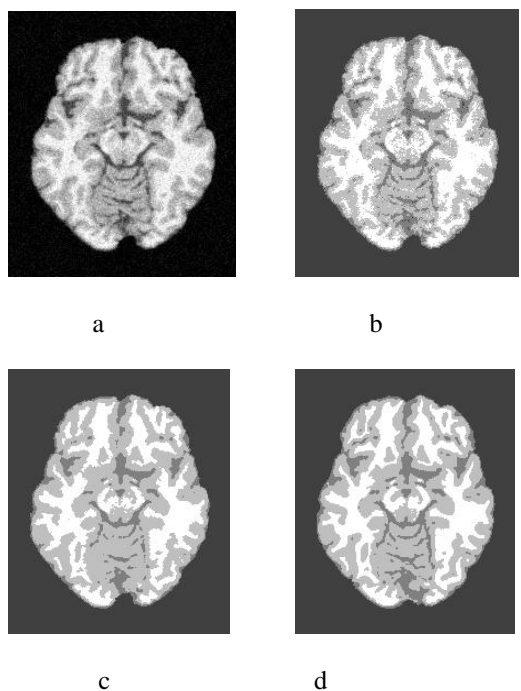

Fig.2. Segmentation results of 15% noise added synthetic brain MRI image. (a) 15% noise added synthetic image. (b) FCM segmentation (c) Proposed approach without denoising (d) Base true.

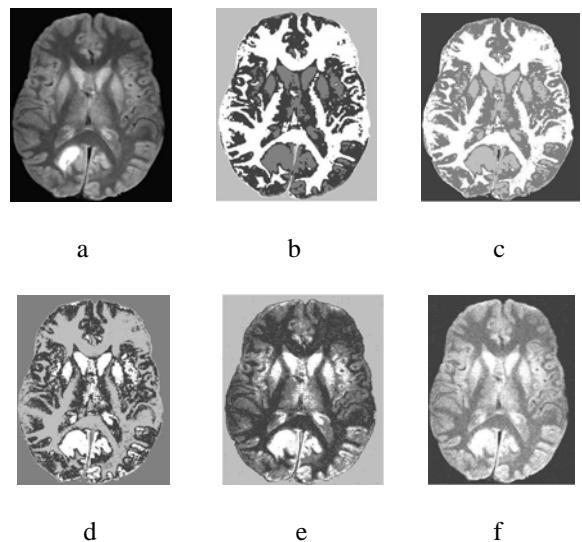

Fig.4. Segmentation results of original brain MRI image. (a) Original brain MRI image with tumor. (b) FCM segmentation with 0% noise (c) with 5% noise (d) with 10 % noise (e) with 15 % noise (f) with 20% noise

The segmentation result of original MRI brain image for proposed algorithm is shown in figure 5.

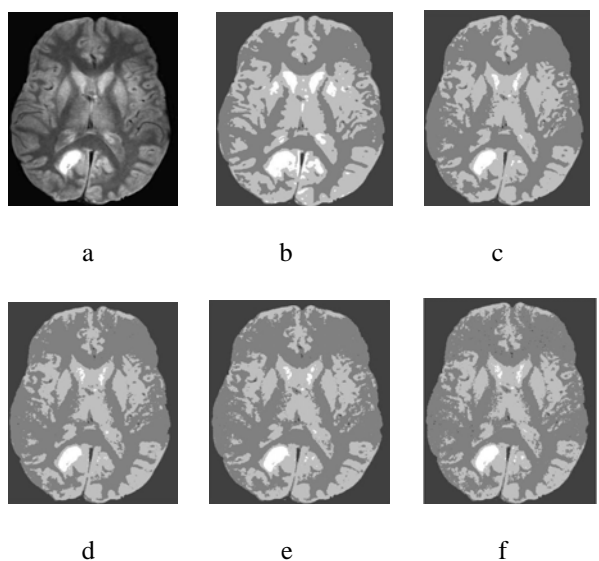

a       b       c

d       e       f

Fig.5. Segmentation results of original brain MRI image. (a)
Original brain MRI image with tumor. (b) ISFCM
segmentation with 0% noise (c) with 5% noise (d) with 10 %
noise (e) with 15 % noise (f) with 20% noise

## V. CONCLUSION

To overcome the noise sensitiveness of conventional FCM
clustering algorithm, this paper presents an improved spatial
fuzzy c mean clustering algorithm(ISFCM) for image
segmentation. The main fact of this algorithm is to incorporate
the spatial neighborhood information into the standard FCM
algorithm by *a priori* probability. The probability can be
automatically decided in the algorithm based on the
membership function of the centre pixel and its neighboring
pixels. The major advantage of this algorithm are its
simplicity, which allows it to run on large datasets. As we
employ a fast FCM algorithm to initialize the ISFCM
algorithm, the algorithm converges after several iterations.
Experimental results show that the proposed method is
effective and more robust to noise and other artifacts than the
conventional FCM algorithm in image segmentation.

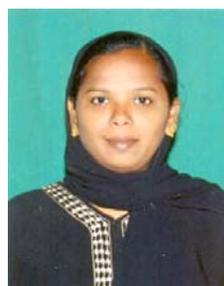

Zulaikha Beevi S. received the B.E.,
degree from the department of Civil and
Transportation Engineering, Institute of
Road and Transport Technology,
TamilNadu, India and M.Tech from the
Department of Computer Science and IT
, Manonmaniam Sundaranar University,
TamilNadu, India in 1992 and 2005,
respectively. She is currently pursuing the
Ph.D. degree, working with
Prof.M.Mohamed Sathik and Prof.K.Senthamarai Kannan. She is
working as Assistant Professor in National College of Engineering ,
Tirunelveli, TamilNadu, India.

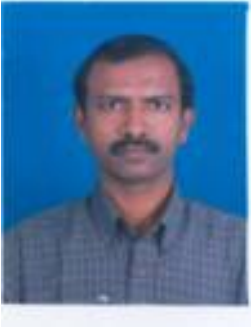

M.Mohamed Sathik completed B.Sc.,and M.Sc.,degrees from Department of Mathematics, M.Phill., from Department of Computer Science, M.Tech from Computer Science and IT ,M.S., from Department of Counseling and Psycho Therapy, and M.B.A degree from reputed institutions. He has two years working experience as a Coordinator for M.Phill Computer Science Program, Directorate of Distance and Continuing Education, M.S.University. He served as Additional Coordinator in Indra Gandhi National Open University for four years. He headed the University Study Center for MCA Week End Courese, Manonmaniam Sundaranar University for 9 years. He has been with the department of Computer Science, Sadakathullah Appa College for 23 years.Now he is working as a Reader for the same department. He works in the field of Image Processing, specializing particularly in medical imaging . Dr.Mohamed Sathik M. guided 30 M.Phil Computer Science Scholors and guiding 14 Ph.D Computer Science Scholor from M.S.University, Tirunelveli, Barathiyar University, Coimbatore and Periyar Maniammai University, Tanjavur. He presented 12 papers in international conferences  in image processing and presented 10 papers in national conferences. He published 3 papers in International Journals and 5 papers in proceedings with ISBN.